\RequirePackage{fix-cm}

\documentclass[twocolumn]{svjour3}
\smartqed
\usepackage{graphicx}
\usepackage{caption}
\usepackage{amsmath}
\usepackage{url}
\begin{document}

\title{NAS-FCOS: 
Efficient Search for Object Detection Architectures
\thanks{
NW, YG, HC contributed to this work equally.}
}

\author{Ning Wang$^{\dag\ddag}$ \and Yang Gao$^{\dag\ddag}$ \and Hao Chen$^\triangle$ \and Peng Wang$^{\dag\ddag}$ \and Zhi Tian$^\triangle$ \and Chunhua Shen$^\diamond$ \and Yanning Zhang$^{\dag\ddag}$
}

\institute{
Peng Wang (
Corresponding author) \at
\email{peng.wang@nwpu.edu.cn} \\
0000-0001-7689-3405 \\ \at
$ ^\dag $ School of Computer Science, Northwestern Polytechnical University, China \\
$ ^\ddag $ National Engineering Lab for Integrated Aero-Space-Ground-Ocean Big Data Application Technology, China \\
$ ^\triangle $ The University of Adelaide, Australia \\
$ ^\diamond $ Monash University, Australia \\
\begin{acknowledgements}
NW, YG, PW and YZ's participation in this work was supported by the National Key R\&D Program of China (No. 2020AAA0106900), and the National Natural Science Foundation of China (No.U19B2037, No.61876152).
\end{acknowledgements}
}

\date{Received: date / Accepted: date}

\maketitle

\begin{abstract}

Neural Architecture Search (NAS) has shown great potential in effectively reducing manual effort in network design by automatically discovering optimal architectures. What is noteworthy is that as of now, object detection is less touched by NAS algorithms despite its significant importance in computer vision. To the best of our knowledge, most of the recent NAS studies on object detection tasks fail to satisfactorily strike a balance between performance and efficiency of the resulting models, let alone the excessive amount of computational resources cost by those algorithms.
Here we propose an efficient method to obtain better object detectors by searching for the feature pyramid network (FPN) as well as the prediction head of a simple anchor-free object detector, namely, FCOS~\cite{tian2019fcos}, using a tailored reinforcement learning paradigm.
With carefully designed search space, search algorithms, and strategies for evaluating network quality, we are able to find top-performing detection architectures within $4$ days using $8$ V100 GPUs.
The discovered architectures surpass state-of-the-art object detection models (such as Faster R-CNN, RetinaNet and, FCOS) by $1.0\%$ to $5.4\%$ points in AP on the COCO dataset, with comparable computation complexity and memory footprint, demonstrating the efficacy of the proposed NAS method for object detection. Code is available at \url{https://github.com/Lausannen/NAS-FCOS}.


\end{abstract}

\section{Introduction}
\label{intro}

In computer vision, object detection is commonly considered as one of the core tasks and the research on it is in full swing.
Generally, the designs of object detection networks are much more complex than those for image classification, because the former need to localize and classify multiple objects in an image simultaneously while the latter only need to output image-level labels.
Due to its complex structure and numerous hyper-parameters, designing effective object detection networks is more challenging and usually needs much manual effort.
In the past few years, state-of-the-art methods for this task are mostly manually designed (such as Faster R-CNN \cite{ren2015faster}, RetinaNet~\cite{lin2017feature} and FCOS~\cite{tian2019fcos}) with impressive performance.

On the other hand, NAS approaches~\cite{ghiasi2019fpn,nekrasov2018fast,zoph2016neural} have been shown
impressive results on automatically discovering top-performing networks in large-scale search spaces, and they can be applied to a variety of computer vision tasks.
Compared to manual designs, NAS frameworks are typically data-driven instead of experience-driven and hence need much less human intervention.

As is summed up in~\cite{elsken2018neural}, the workflow of a general NAS method can be divided into the following three processes:

\begin{itemize}
\item[1)] sampling architecture from a search space following some search strategies;

\item[2)] evaluating the performance of the sampled architecture;

\item[3)] updating parameters of the proxy networks by optimizing the against some performance measures.
\end{itemize}

Search efficiency is one of the main problems prohibiting NAS from being used in more realistic applications. The performance evaluation process is the most time-consuming part because it involves a full training procedure of a neural network. In order to obtain accurate metrics, the ideal state is to fully train the sampled architecture using standard datasets and training strategies in the corresponding tasks. But for tasks like object detection, complete training of a single architecture already requires several hours and even days on multiple GPUs.
Therefore, the full training in the search phase is not realistic under the current computation constraints.

To reduce the evaluation time, in practice a proxy task is often used as a lower cost substitution. In the proxy task, the input size, network parameters, and training iterations are often scaled down to speed up the evaluation. However, there is often a performance gap for samples between the proxy tasks and target tasks, which makes the evaluation process biased. How to elaborately design proxy tasks that are both accurate and efficient for specific problems is a key issue to the success of NAS.

Another solution to improve search efficiency is to construct a supernet that covers the complete search space and train candidate architectures with shared parameters~\cite{liu2018darts,pham2018enas}.
However, this solution often leads to significantly increased memory consumption and restricts itself to small-to-moderate sized search spaces.
Since the model weights cannot be decoupled directly, the performance gap still exists. So far, no research has shown that one search strategy is necessarily better than another, which is prompting researchers to continuously expand their ideas.

Since efficient and accurate NAS approaches for object detection are less studied, we present a fast and memory-saving NAS method for object detection networks, which is capable of 
discovering top-performing architectures within significantly reduced search time.
Our overall detection architecture is based on FCOS~\cite{tian2019fcos},
a simple anchor-free one-stage object detection framework, in which the feature pyramid network and prediction head are searched using our proposed NAS method. Here we want to clarify that with the same architecture, anchor-free detectors can save time by cancelling the processing time of anchor-box matching in anchor-based detectors like RetinaNet. Considering that thousands of models need to be evaluated, the time saved by a single model is magnified.

Our main contributions are summarized as follows:
\begin{itemize}
    \item[1)] In this work, we propose a fast and memory-efficient NAS method for
    searching both FPN and head architecture of object detectors with carefully designed proxy tasks, search space, and evaluation strategies, which is able to evaluate more than $3,000$ top-performing architectures using $28$ GPU-days only.
    
    Specifically, this high efficiency is enabled with the following designs:
    
    $-$ Developing a fast proxy task training scheme by skipping the backbone finetuning stage;
    
    $-$ Adapting progressive search strategy to reduce time cost taken by the extended search space;
    
    $-$ Using a more discriminative criterion for evaluation of searched architectures.
    
    $-$ Employing an efficient anchor-free one-stage detection framework with simple post-processing;

    \item[2)]
    Using NAS, we discuss the workload relationship between FPN and head, exploring the importance of weight sharing at both head-level and branch-level which is defined in Section $3.2.3$.
    
    \item[3)] 
    We show that the overall structure of NAS-FCOS is general and flexible in that it can be equipped with various backbones including MobileNetV$2$, ResNet-$50$, ResNet-$101$ and ResNeXt-$101$, and surpasses st\-a\-t\-e-o\-f-the-art object detection algorithms using comparable computation complexity and memory footprint. More specifically, our models can improve the AP by $1.0\% \sim5.4\%$ points on all the above models comparing to their FCOS counterparts.

\end{itemize}

Note that compared with the conference version, this version has been considerably extended as follows:
\begin{itemize}

\item[1)] Firstly, we now use
a carefully-selected
COCO subset to further refine proxy
tasks. The original PASCAL VOC can be used to investigate the transfer
capacity of the searched architecture while the COCO subset is for better
fitting the complete MS-COCO dataset during the evaluation process.
Compared with the VOC-based proxy task, the COCO-based one
further improves the detection performance by about $0.5\%$ AP under the
comparable computation complexity.

\item[2)] Secondly, a more comprehensive weight-sharing mechanism is
explored at the branch-level. With the help of NAS, we study the trend of
feature sharing between the classification branch and the regression branch.
Our result shows that a single feature may not be suitable for
different tasks.

\item[3)] Thirdly, the proposed method is not specifically designed for anchor-free models. Actually the searched structure can be directly transferred to anchor-based models such as RetinaNet~\cite{lin2017focal}. By combining it with the RetinaNet detector, our NAS found architecture \textit{significantly} outperforms the original RetinaNet by  $3.4$ points in AP using ResNet-$50$ backbone (AP: $39.5\%$ vs.\  $36.1\%$), showing that anchor-based models can also benefit from our method.

\end{itemize}

\section{Related Work}
\label{sec:1}

\subsection{Object Detection}
\label{sec:2}
Generally, object detectors consist of four parts including backbones, necks, region proposal networks (RPNs), and heads. Note that backbones are used to extract features from input images, and necks (e.g, FPN) are responsible for blending multi-level features. Region proposal networks generate prediction candidates on extracted features and pass them to heads to further refine localization and classify the specific type. Depending on whether there is an RPN, deep learning frameworks for object detection can be roughly categorized into two types:
one-stage detectors~\cite{lin2017focal} and two-stage detectors~\cite{he2017mask,ren2015faster}.

Specifically, two-stage detection frameworks firstly generate class-independent region proposals using an RPN
and then classify and refine them using extra detection heads. The typical detection paradigm is the R-CNN series. R-CNN~\cite{rgb2014rcnn} firstly introduces CNNs to object detection and achieves impressive performance. Selective Search~\cite{us2013selective} is applied to generate proposals and each region is then classified by SVM. Due to repeated feature calculations, R-CNN is limited by its slow inference time. Following R-CNN, Fast R-CNN~\cite{rgb2015fast-rcnn} is designed to solve this problem by sharing the computation of features between proposals. The proposed RPN enables end-to-end training of the two-stage detection framework and further improves accuracy in Faster R-CNN~\cite{ren2015faster}. By adding extra branches, Mask R-CNN~\cite{he2017mask} unifies more visual tasks (e.g. instance segmentation, key-point detection) into a single framework. Despite achieving top performance, the two-stage methods have two noticeable drawbacks: 
they are computationally expensive and have many hyper-parameters that need to be tuned to fit a specific dataset.

In comparison, the structures of one-stage detectors are much simpler.
They directly predict object categories and bounding boxes at each location of feature maps generated by a single CNN backbone. YOLO~\cite{jr2016yolov1} and YOLOv$2$~\cite{jr2017yolov2} directly conduct classification and regression on the features extracted from inputs while ignoring the RPNs to further boost the inference. SSD~\cite{liu2016ssd} introduces a set of bounding boxes from several feature maps with different scales to make up for the AP loss due to the lack of RPNs to some extent. YOLOv$3$, v$4$, v$5$~\cite{yolov3,bochkovskiy2020yolov4} further draw lessons from many detection techniques to greatly improve accuracy without loss of speed. And Efficient-Det~\cite{tan2020efficientdet} optimizes the structure and hyper-parameters as a whole, achieving empirical detection performance.

Note that most state-of-the-art object detectors (including both one-stage detectors and two-stage detectors) make predictions based on anchor boxes of different scales and aspect ratios at each convolutional feature map location. 
However, the usage of anchor boxes may lead to a high imbalance between object and non-object examples and introduce extra hyper-parameters. 
More recently, anchor-free one-stage detectors~\cite{kong2019foveabox,law2018cornernet,tian2019fcos,zhou2019objects,zhu2019fsaf,Yang2020DenseRep} have attracted increasing research interests, due to their simple fully convolutional architectures and reduced consumption of computational resources.

\subsection{Neural Architecture Search}
\label{sec:3}
NAS is firstly introduced by~\cite{zoph2016neural} and quickly attracts the attention of researchers. Typically, NAS can be divided into three parts, including search space, search algorithms, and evaluation indicators. The search space defines model components of different granularity and supports searching for different operations and connections. Since search space often contains tens of thousands of candidate structures, it is not realistic to traverse one by one. The search algorithms tend to find a relatively optimal sub-structure in space to meet the requirements of accuracy and efficiency. And evaluation indicators distinguish between different structures and further guide the search process.

In~\cite{chen2020mnasfpn,zoph2016neural,Zoph2018learning}, researchers adopt RL to drive the search process. The RL system is learning what to do - how to map situations to actions - to maximize a numerical reward signal. One can identify three main subelements of the RL system: the policy, the reward signal and the model of the environment. The policy defines a mapping from states of the environment to actions while the reward signal will give feedback on an action. The above both must interact in the environment. To some extent, we can map the components of NAS to the elements of RL respectively. The evaluator plays the role of the environment, and it gives specific indicators as the reward. The policy will be instantiated by a controller (usually using an LSTM). RL-based NAS also follows the general workflow which has been described in the introduction part. And the key to this approach is to have the controller learn the mapping between the encoded structure to the evaluation indicator.

NAS is notoriously time-consuming. We have seen great improvements from $24,000$ GPU-days~\cite{zoph2016neural} to $0.2$ GPU-day~\cite{zhou2019bayesnas}. The trick is to first construct a supernet containing the complete search space and train the candidates all at once with bi-level optimization and efficient weight sharing~\cite{liu2019auto,liu2018darts}. But the large memory allocation and difficulties in approximated optimization prohibit the search for more complex structures.

Recently researchers~\cite{cai2018proxylessnas,guo2019single,stamoulis2019single} propose to apply single-path training to reduce the bias introduced by approximation and model simplification of the supernet. DetNAS~\cite{chen2019detnas}
follows this idea to search for an efficient object detection architecture.
One limitation of the single-path approach is that the search space is restricted to a sequential structure. Single-path sampling and straight through the estimate of the weight gradients introduce large variance to the optimization process and prohibit the search for more complex structures under this framework. Within this very simple search space, NAS algorithms can only make trivial decisions like kernel sizes for manually designed modules. 

Object detection models are different from single-path image classification networks in their way of merging multi-level features and distributing the task to parallel prediction heads. The FPN series~\cite{ghiasi2019fpn,Alexander2019panoptic,lin2017feature,Liu2019AnEnd,zhao2019pyramid,tan2020efficientdet} which are designed to handle this job play an important role in modern object detection models. NAS-FPN~\cite{ghiasi2019fpn} targets searching for an FPN alternative based on a one-stage framework RetinaNet~\cite{lin2017focal}. Feature pyramid architectures are sampled with a recurrent neural network (RNN) controller. The RNN controller is trained with RL. However, the search is very time-consuming even though a proxy task with ResNet-$10$ backbone is trained to evaluate each architecture.

As this line of research (\cite{chen2019detnas,ghiasi2019fpn,guo2020hit,jiang2020sp,chen2020mnasfpn,du2020spinenet,peng2019efficient,xiong2020mobiledets,yao2019sm,liang2019computation,zhong2020representation,xu2019auto} and ours) focuses on the detection framework, we discuss the differences as follows.
{\
SpineNet~\cite{du2020spinenet} aims to search for the designs of better backbones (including the FPN structure), while DetNAS~\cite{chen2019detnas}, SP-NAS~\cite{jiang2020sp} and CR-NAS~\cite{liang2019computation} focus on the individual backbone structure. In contrast, NAS-FPN~\cite{ghiasi2019fpn} searches the FPN structure and MnasFPN~\cite{chen2020mnasfpn}, FAD~\cite{zhong2020representation} explore the search space on the head structure.
Unlike the above model, Auto-FPN~\cite{xu2019auto} has the same search space (both FPN and head structure) as our model, but it follows the idea of DARTS~\cite{liu2018darts} and uses the gradient-based search strategy.
Compared with Auto-FPN, our RL-based method could be more flexible to design and easy to extend the search space.
Note that both Hit-Detector~\cite{guo2020hit} and SM-NAS~\cite{yao2019sm} claim their frameworks as efficient methods to discover architectures of all components (i.e., backbone, neck, and head) for detection task, the design of our search space for FPN structure is much more complicated than those of theirs, which leads to better results on target dataset. Furthermore, MobileDets~\cite{xiong2020mobiledets} only pays attention to build blocks in detection models on mobile devices, and NATS~\cite{peng2019efficient} proposes an NAS strategy in channel-level instead of path-level while the latter is our main research theme.
}

To speed up the reward evaluation of RL-based NAS, the work of~\cite{nekrasov2018fast} proposes to use progressive tasks and other training acceleration methods. By caching the encoder features, they can train semantic segmentation decoders with very large batch sizes very efficiently. In the sequel of this paper, we refer to this technique as a fast decoder adaptation. However, directly applying this technique to object detection tasks does not enjoy a similar speed boost, because they are either not in using a fully-convolutional model~\cite{lin2017feature} or require complicated post-processing that is not scalable with the batch size~\cite{lin2017focal}.

To reduce the post-processing overhead, we resort to a recently introduced anchor-free one-stage framework, namely, FCOS~\cite{tian2019fcos}, which significantly improve the search efficiency by canceling the processing time of anchor-box matching in RetinaNet.

Compared to its anchor-based counterpart, FCOS significantly reduces the training memory footprint while being able to improve the performance. Note that although we adopt FCOS to further improve search efficiency, the searched networks can be equipped with both anchor-free and anchor-based detectors.

\section{Our Approach}
\label{sec:4}

In our work, we search for anchor-free fully convolutional detection models with fast decoder adaptation. Thus, NAS methods can be easily applied.

\subsection{Problem Formulation}
\label{sec:5}
We base our search algorithm upon a one-stage framework FCOS due to its simplicity. Our training tuples $\{(\mathbf x, Y)\}$ consist of input image tensors $\mathbf x$ of size $(3\times H\times W)$ and FCOS output targets $Y$ in a pyramid representation, which is a list of tensors $\mathbf y_l$ each of size $((K+4+1)\times H_l\times W_l)$ where $H_l\times W_l$ is feature map size on level $p$ of the pyramid. $(K+4+1)$ is the output channels of FCOS, the three terms are length-$K$ one-hot classification labels, $4$ bounding box regression targets and $1$ centerness factor respectively.

The network $g: \mathbf x\rightarrow \hat{Y}$ in original FCOS consists of three parts, a backbone $b$, FPN $f$ and multi-level subnets we call prediction heads $h$ in this paper. First backbone $b: \mathbf x\rightarrow C$ maps the input tensor to a set of intermediate-leveled features $C = \{\mathbf c_3, \mathbf c_4, \mathbf c_5\}$, with resolution $(H_i\times W_i) = (H/2^i \times W/2^i)$. Then FPN $f: C\rightarrow P$ maps the features to a feature pyramid $P=\{\mathbf p_3, \mathbf p_4, \mathbf p_5, \mathbf p_6, \mathbf p_7\}$. Then the prediction head $h: \mathbf p\rightarrow \mathbf y$ is applied to each level of $P$ and the result is collected to create the final prediction. To avoid overfitting, same $h$ is often applied to all instances in $P$.

Since objects of different scales require different effective receptive fields, the mechanism to select and merge intermediate-leveled features $C$ is particularly important in object detection network design. Thus, most researches~\cite{liu2016ssd,ren2015faster} are carried out on designing $f$ and $h$ while using widely-adopted backbone structures such as ResNet~\cite{he2016identity}. Following this principle, our search goal is to decide when to choose which features from $C$ and how to merge them.

To improve the efficiency, we reuse the parameters in $b$ pretrained on target dataset and search for the optimal structures after that. For the convenience of the following statement, we call the network components to search for, namely $f$ and $h$, together the decoder structure for the objection detection network.

$f$ and $h$ take care of different parts of the detection job. $f$ extracts features targeting different object scales in the pyramid representations $P$, while $h$ is a unified mapping applied to each feature in $P$ to avoid overfitting. In practice, people seldom discuss the possibility of using a more diversified $f$ to extract features at different levels or how many layers in $h$ need to be shared across the levels. In this work, we use NAS as an automatic method to test these possibilities.

\subsection{Search Space}
\label{sec:6}

\begin{figure*}[thb!]
	\centering
	\includegraphics[width = 0.95\linewidth]{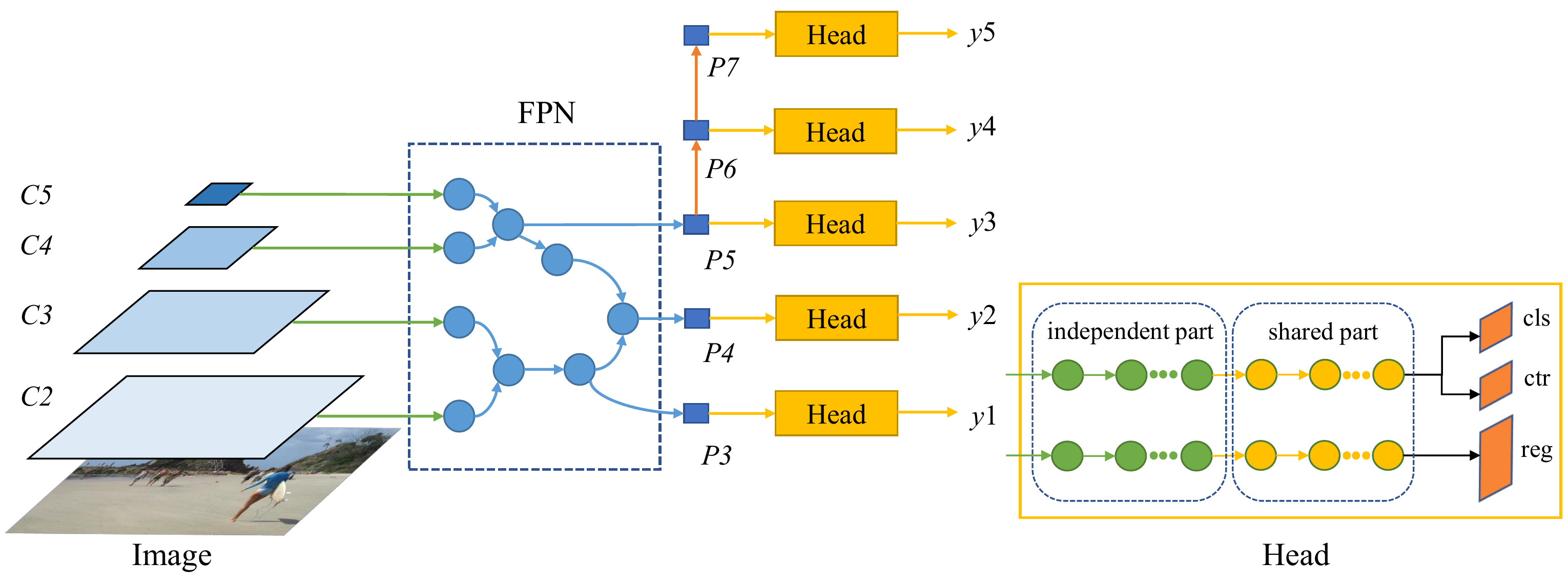}
	\caption{A conceptual example of our NAS-FCOS decoder. It consists of two sub networks, an FPN $f$ and a set of prediction heads $h$ which have shared structures. One notable difference with other FPN-based one-stage detectors is that our heads have partially shared weights. Only the last several layers of the predictions heads (not marked in green) are tied by their weights at head-level. And branch-level weight sharing is also explored in these layers. The number of layers to share is decided automatically by the search algorithm. Note that both FPN and head are in our actual search space; and have more layers than shown in this figure. Here the figure is for illustration only.
\label{fig:main}}
\end{figure*}

We mainly focus on the decoder part in the detector. The decoder configuration can be represented by a sequence with three components, FPN configuration, head configuration and weight sharing stages. We provide detailed descriptions to each of them in the following sections. The complete diagram of our decoder structure is shown in Fig.~\ref{fig:main}.

Considering the different functions of $f$ and $h$, we apply two search space design respectively. Given the particularity of FPN structure, a basic block with new overall connection and $f$'s output design is built for it. For simplicity, sequential space is applied for $h$ part.

\subsubsection{FPN Search Space}
\label{sec:7}

As mentioned above, the FPN $f$ maps the convolutional features $C$ to $P$. 
First, we initialize the sampling pool as $X_0 = C$. To construct one basic block, we first choose two layers $\mathbf x_1$, $\mathbf x_2$ from the sampling pool $X$ at \texttt{id1}, \texttt{id2}, then two operations \texttt{op1}, \texttt{op2} are applied to each of them and an aggregation operation \texttt{agg} merges the two output into one feature. To build a deep FPN space, we apply multiple basic blocks with their outputs added to the sampling pool. Our basic block $bb_t: X_{t-1}\rightarrow X_t$ at time step $t$ transforms the sampling pool $X_{t-1}$ to $X_t = X_{t-1}\cup \{\mathbf{x}_t\}$, where $\mathbf{x}_t$ is the output of $bb_t$.
Our FPN is defined by applying $7$ basic blocks to the sampling pool, $f:= bb_1^f\circ bb_2^f \circ \cdots \circ bb_7^f$. Note that each basic block has its own topology, allowing to search in a larger space. To yield pyramid features $P$, we collect the last three basic block outputs $\{\mathbf x_5, \mathbf x_6, \mathbf x_7\}$ as $\{\mathbf p_3, \mathbf p_4, \mathbf p_5\}$. Consistent with FCOS, $\mathbf{p}_6$ and $\mathbf{p}_7$ are obtained via a $3\times3$ stride-$2$ convolution on $\mathbf{p}_5$ and $\mathbf{p}_6$ respectively.

To allow shared information across all layers, we use a simple rule to create global features. If there is some dangling layer $\mathbf x_t$ which is not sampled by later blocks $\{bb_i^f|i > t\}$ nor belongs to the last three layers $t < 5$, we use element-wise add to merge it to all output features
\begin{equation}
    \mathbf{p}^*_i = \mathbf p_i + \mathbf x_t, \,\, i\in\{3, 4, 5\}.
\end{equation}

Same as the aggregation operations, if the features have different resolution, the smaller one is upsampled with bilinear interpolation.

The candidate operations are listed in Table~\ref{table:unary}. We include only separable convolutions so that the decoder can be efficient. In order to enable the decoder to apply convolutional filters on irregular grids, here we have also included deformable $3\times3$ convolutions~\cite{zhu2018deformable}.
For the aggregation operations, we include element-wise sum and
concatenation followed by a $1\times1$ convolution.

\begin{table}[h!]
\label{tab:1}
	\begin{center}
	    \begin{tabular}{c|c}
            \hline 
            ID & Description \\
			\hline
			0 & separable conv $3\times3$ \\
			\hline
			1 & separable conv $3\times3$ with dilation rate $3$\\
			\hline
			2 & separable conv $5\times5$ with dilation rate $6$\\
			\hline
			3 & skip-connection\\
			\hline
			4 & deformable $3\times3$ convolution\\
            \hline
		\end{tabular}
		\smallskip
		\caption{Unary operations used in the search process.
			\label{table:unary}}
	\end{center}
\end{table}

\subsubsection{Prediction Head Search Space}
\label{sec:8}

\def\x{{$\times$}}

Prediction head $h$ maps each feature in the pyramid $P$ to the output of corresponding $\mathbf y$, which in FCOS and RetinaNet, consists of four $3\times3$ convolutions. To explore the potential of the head, we therefore extend a sequential search space for its generation. Specifically, our head is defined as a sequence of six basic operations. Compared with candidate operations in the FPN structures, the head search space has two slight differences. First, we add standard convolution modules (including conv$1$\x$1$ and conv$3$\x$3$) to the head sampling pool for better comparison. Second, we follow the design of FCOS by replacing all the Batch Normalization (BN) layers to Group Normalization (GN)~\cite{wu2018group} in the operations sampling pool of head, considering that head needs to share weights between different levels, which causes BN invalid.
The final output of head is the output of the last (sixth) layer.

\subsubsection{Searching for Head Weight Sharing}
\label{sec:9}

Modern detectors tend to share weights between heads in different levels while keeping branches own independent parameters inside one head.
To add even more flexibility and understand the effect of weight sharing or independent in prediction heads, we further add two indexes $i$ and $j$ as the locations where the prediction head starts or ends to share weights.

Firstly, we consider extending the independent part of the heads into FPN, and use the shared part as head. Since index $i$ determines the ratio of weight-sharing between heads, we can further balance the workload for each individual FPN part to facilitate the extraction of level-specific features and the prediction head shared by all levels. We call the above head-level weight sharing and index $i$ is responsible for this exploration. Specifically, for every layer before stage $i$, the head $h$ will create independent set of weights for each FPN output level, otherwise, it will use the global weights for sharing purpose. 

Secondly, the effect of unequal features between different branches remains to be discussed. In FCOS, independent branches are adapted for both classification and regression.
What we call the branch-level weight sharing can be used to study the importance of re-using features in the above two branches and handed over to $j$ at the implementation level. Note that we explore this point in the shared part of the head whose layers are defined after index $i$. Concretely speaking, classification and regression branches share the weights before $j$, and then use individual weights for each other. For convenience, we will take the expression of 'head-level' and 'branch-level' sharing in the next following sections.

\subsection{Search Strategy}
\label{search strategy}
RL based strategy is applied to the search process. We rely on an LSTM-based controller to predict the full configuration.
We consider using a progressive search strategy rather than the joint search for both FPN structure and prediction head part, since the former requires less computing resources and time cost than the latter. More specifically, the progressive search firstly search $f$ with the manually designed $h$ and then search $h$ with the searched $f$. The training dataset is randomly split into a meta-train $D_t$ and meta-val $D_v$ subset. To speed up the training, we fix the backbone network and cache the pre-computed backbone output $C$. This makes our single architecture training cost independent from the depth of backbone network. Taking this advantage, we can apply much more complex backbone structures and utilize high quality multilevel features as our decoder's input. 
We find that the process of backbone finetuning can be skipped if the cached features are powerful enough. Speedup techniques such as Polyak weight averaging are also applied during the training.

The most widely used detection metric is average precision (AP). However, due to the difficulty of object detection task, at the early stages, we found that AP is too low to tell the good architectures from the bad ones, which makes the controller take much more time to converge. For the purpose of making the architecture evaluation process easier even at the early stages of the training, we therefore use the sum of negative loss as the reward instead of average precision:
\begin{equation}
\begin{split}
    R(a) = - \sum_{(x, Y)\in D_v}(L_{cls}(x, Y|a) + L_{reg}(x, Y|a) \\
    + L_{ctr}(x, Y|a))
\end{split}
\label{reward}
\end{equation}
where $L_{cls}$, $L_{reg}$, $L_{ctr}$ are the three loss terms in FCOS. Gradient of the controller is estimated via proximal policy optimization (PPO)~\cite{schulman2017proximal}.

\section{Experiments}
\label{sec:11}

\subsection{Implementation Details}
\label{sec:12}

\subsubsection{Searching Phase}
\label{sec:13}

We design a fast proxy task for evaluating the decoder architectures sampled in the searching phase.

The PASCAL VOC and a carefully-selected MS-COCO subset are both explored as the proxy task. The former is for investigating the transfer capacity of the searched structures and the latter is for better fitting the complete MS-COCO dataset during the evaluation process.

The PASCAL VOC dataset contains $5715$ training images with object bounding box annotations of $20$ classes. In this work, the VOC training set is randomly split into a meta-train set with $4,000$ images and a meta-val set with $1715$ images.

As for the carefully-selected MS-COCO dataset, we first randomly sample the same number of pictures as VOC. However, more categories and smaller detection targets make candidate networks training more difficult. In order to obtain a better proxy dataset on COCO, three factors are considered: 
instance number, average area, and average ratio of bounding-box (bbox) to image area per category. Note that the average ratio is obtained through dividing the area of the bounding box by the original images. Based on the above factors, we have counted the information of 80 categories in COCO. The interval covered by the indicator value is calculated and then divided into $5$ segments. We sample $4$ categories in each segment to ensure that it is as close as possible to the data distribution of the original COCO. $20$ categories in total are shown in Table~\ref{table:dataset} as the proxy dataset.

\begin{table}[th!]
\centering
\scalebox{0.80}{
    \begin{tabular}{l|c|c|c|c}
    \hline\noalign{\smallskip}
    ID & Name & Instance & Avg Area & Avg Ratio (bbox/image)(\%) \\
    \noalign{\smallskip}\hline\noalign{\smallskip}
    1 & bottle & $24342$ & $4451.3$ & $1.6$\\
    \noalign{\smallskip}\hline\noalign{\smallskip}
    2 & book & $24715$ & $5932.8$ & $2.2$ \\
    \noalign{\smallskip}\hline\noalign{\smallskip}
    3 & skateboard & $5543$ & $9310.9$ & $3.4$ \\
    \noalign{\smallskip}\hline\noalign{\smallskip}
    4 & bird & $10806$ & $13129.2$ & $4.8$ \\
    \noalign{\smallskip}\hline\noalign{\smallskip}
    5 & sink & $3587$ & $16069.4$ & $5.8$ \\
    \noalign{\smallskip}\hline\noalign{\smallskip}
    6 & banana & $9458$ & $18011.3$ & $6.4$ \\
    \noalign{\smallskip}\hline\noalign{\smallskip}
    7 & person & $262465$ & $22284.6$ & $8.1 $ \\
    \noalign{\smallskip}\hline\noalign{\smallskip}
    8 & bench & $9838$ & $23271.9$ & $8.4 $ \\
    \noalign{\smallskip}\hline\noalign{\smallskip}
    9 & suitcase & $6192$ & $26255.5$ & $9.6 $ \\
    \noalign{\smallskip}\hline\noalign{\smallskip}
    10 & cake & $6353$ & $30014.3$	& $10.4$ \\ 
    \noalign{\smallskip}\hline\noalign{\smallskip}
    11 & fire hydrant & $1865$ & $31667.1$ & $11.8$ \\
    \noalign{\smallskip}\hline\noalign{\smallskip}
    12 & motorcycle & $8725$ & $35192.1$ & $12.5$ \\
    \noalign{\smallskip}\hline\noalign{\smallskip}
    13 & laptop & $4970$ & $39638.3$ & $14.6$ \\
    \noalign{\smallskip}\hline\noalign{\smallskip}
    14 & oven & $3334$ & $47937.5$ & $17.0$ \\
    \noalign{\smallskip}\hline\noalign{\smallskip}
    15 & bus & $6069$ & $52573.3$ & $18.6$ \\
    \noalign{\smallskip}\hline\noalign{\smallskip}
    16 & giraffe & $5131$ & $57309.6$ & $21.0$ \\
    \noalign{\smallskip}\hline\noalign{\smallskip}
    17 & pizza & $5821$ & $63539.5$ & $21.8$ \\
    \noalign{\smallskip}\hline\noalign{\smallskip}
    18 & couch & $5779$ & $62755.8$ & $23.2$ \\
    \noalign{\smallskip}\hline\noalign{\smallskip}
    19 & train & $4571$ & $79644.8$ & $28.8$ \\
    \noalign{\smallskip}\hline\noalign{\smallskip}
    20 & dining table & $15714$ & $102777.2$ & $35.8$ \\
    \noalign{\smallskip}\hline
    \end{tabular}
}
    \smallskip
    \caption{20 categories which are sampled from COCO to construct the proxy subset.
	\label{table:dataset}}
\end{table}

For each sampled architecture, we train it on meta-train and compute the reward~\eqref{reward} on meta-val.
Input images of two proxy datasets are both resized to short size $384$ and then randomly cropped to $384\times384$. Target object sizes of interest are scaled correspondingly. We use Adam optimizer with learning rate $8$e$-4$ and batch size $200$. Polyak averaging is applied with the decay rates of $0.9$. The decoder is evaluated after $300$ iterations.

As we use fast decoder adaptation, the backbone features are fixed and cached during the search phase.
To enhance the cached backbone features, we first initialize them with pre-trained weights provided by the open-source implementation of FCOS
and then finetune on the proxy dataset using the training strategies of FCOS. Note that the above finetuning process is only performed once at the beginning of the search phase.  

A progressive strategy is used for the search of $f$ and $h$. We first search for the FPN part and retain the original head. All operations in the FPN structure have $64$ output channels. The decoder inputs $C$ are resized to fit the output channel width of FPN via $1\times1$ convolutions. After this step, a searched FPN structure is fixed and the second stage searching for the head will be started based on it. Most parameters for the searching head are identical to those for searching FPN structure, with the exception that the output channel width is adjusted from $64$ to $128$ to deliver more information.

\begin{figure}[t!]
\centering
\includegraphics[width=0.48\textwidth]{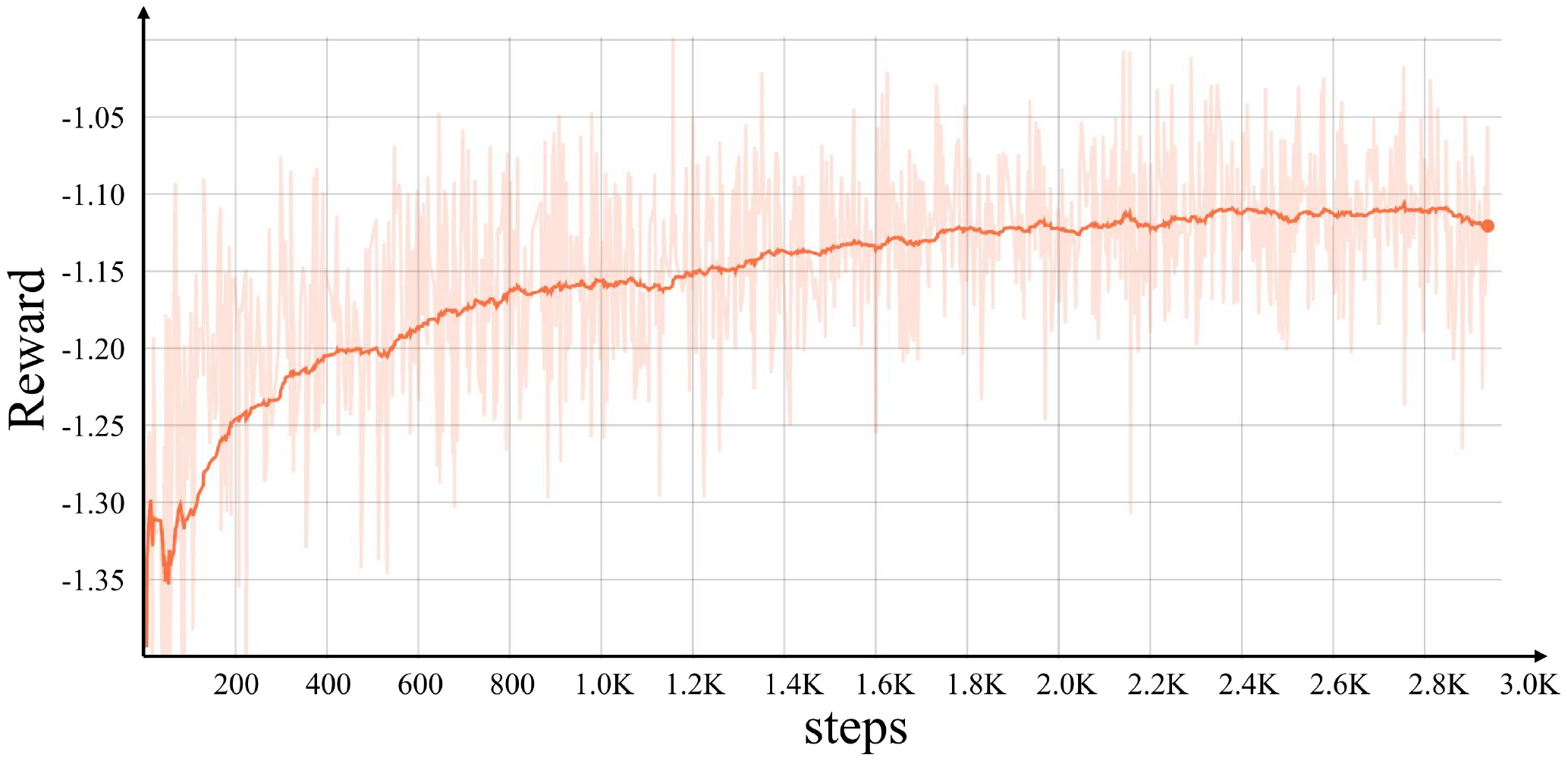}
\caption{Performance of reward during the VOC proxy task, which has been growing throughout the process, indicating that the model of reinforcement learning works.}
\label{fig:reward}
\end{figure}

\begin{figure}[t!]
\centering
\includegraphics[width=0.48\textwidth]{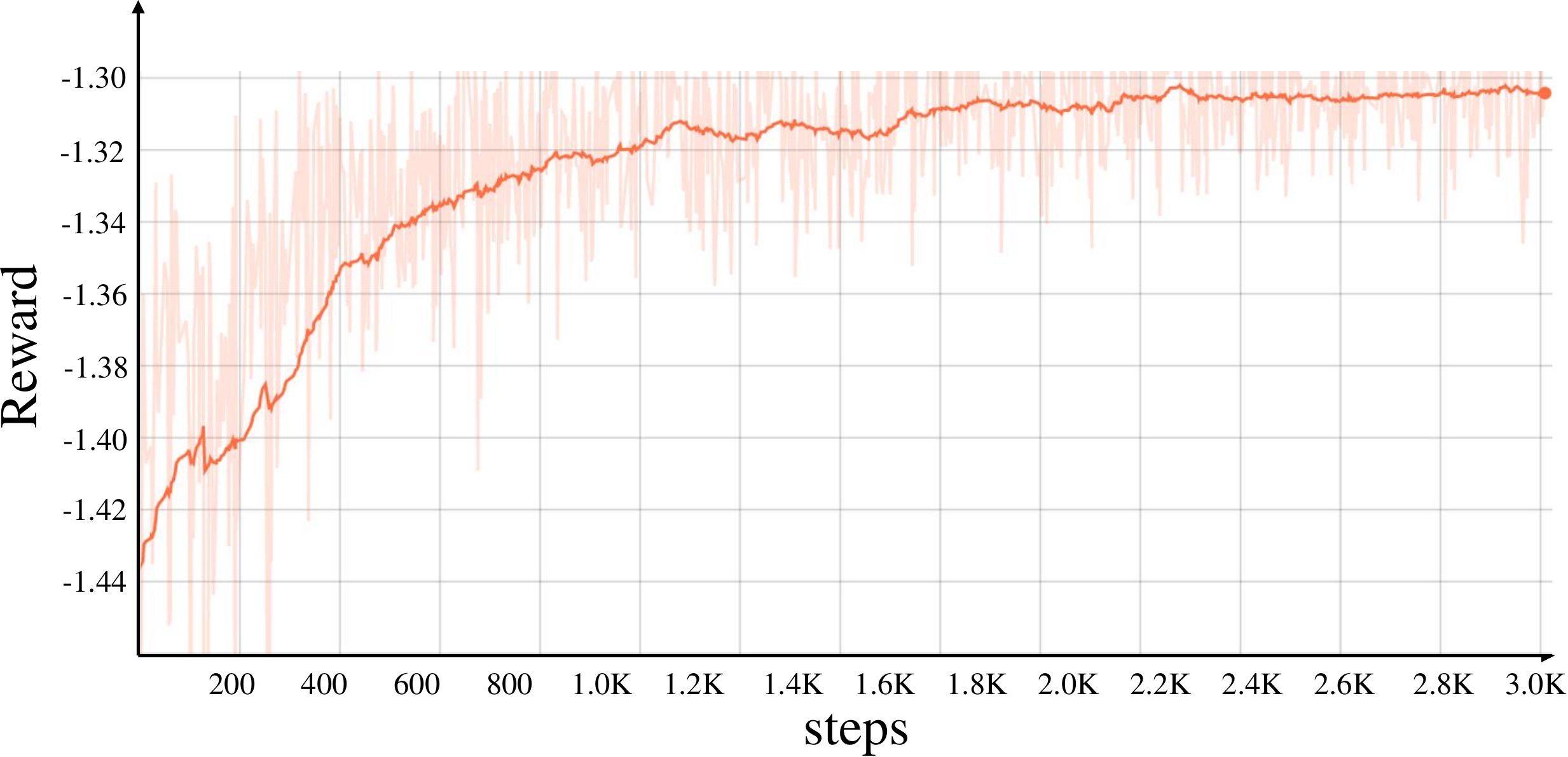}
\caption{Performance of reward during the COCO proxy task. Compared with Fig~\ref{fig:reward}, the reward is lower overall due to the complex COCO subset.}
\label{fig:coco_reward}
\end{figure}

For the FPN search part, the controller model nearly converged after searching over $2.8$K architectures on the VOC and COCO proxy task as shown in Fig.~\ref{fig:reward} and Fig.~\ref{fig:coco_reward} respectively. Then, the top-$20$ best performing architectures on the proxy task are selected for the next full training phase.
For the head search part, we choose the best searched FPN among the top-$20$ architectures and pre-fetch its features. It takes about $600$ rounds for the controller to nearly converge, which is much faster than that for searching FPN architectures. After that, we select for full training the top-$10$ heads that achieve the best performance on the proxy task. In total, the whole search phase can be finished within $4$ days using $8$ V100 GPUs.

\begin{table*}[th!]
\centering
\scalebox{0.9}{
    \begin{tabular}{ll|cc|cc}
    \hline\noalign{\smallskip}
    Decoder & Backbone & FLOPs (G) & Params (M) & AP\\
    \noalign{\smallskip}\hline\noalign{\smallskip}
    FPN-RetinaNet @$256$ & MobileNetV$2$ & $133.4$ & $11.3$ & $30.8$\\
    FPN-FCOS @$256$ & MobileNetV$2$ & $105.4$ & $9.8$ & $31.2$\\
    NAS-FCOS (VOC) @$128$ & MobileNetV$2$ & $\mathbf{39.3}$ & $\mathbf{5.9}$ & $32.0$\\
    NAS-FCOS (VOC) @$128$-$256$ & MobileNetV$2$ & $95.6$ & $9.9$ & $33.8$\\
    NAS-FCOS (VOC) @$256$ & MobileNetV$2$ & $121.8$ & $16.1$ & $34.7$\\
    NAS-FCOS (COCO) @$256$ & MobileNetV$2$ & $102.9$ & $15.0$ & $\mathbf{36.2}$\\
    \noalign{\smallskip}\hline\noalign{\smallskip}
    FPN-RetinaNet @$256$ & R-$50$ & $198.0$ & $33.6$ & $36.1$\\
    FPN-FCOS @$256$ & R-$50$ & $169.9$ & $32.0$ & $37.4$\\
    NAS-FCOS (VOC) @$128$ & R-$50$ & $\mathbf{104.0}$ & $\mathbf{27.8}$ & $37.9$\\
    NAS-FCOS (VOC) @$128$-$256$ & R-$50$ & $160.4$ & $31.8$ & $39.1$\\
    NAS-FCOS (VOC) @$256$ & R-$50$ & $189.6$ & $38.4$ & $39.8$\\
    NAS-FCOS (COCO) @$256$ & R-$50$ & $167.5$ & $37.3$ & $\mathbf{40.3}$\\
    \noalign{\smallskip}\hline\noalign{\smallskip}
    FPN-RetinaNet @$256$ & R-$101$ & $262.4$ & $52.5$ & $39.1$\\
    FPN-FCOS @$256$ & R-$101$ & $\mathbf{234.3}$ & $\mathbf{50.9}$ & $41. 5$\\
    NAS-FCOS (VOC) @$256$ & R-$101$ & $254.0$ & $57.3$ & $43.0$\\
    NAS-FCOS (COCO) @$256$ & R-$101$ & $231.9$ & $56.2$ & $43.4$\\
    NAS-FCOS (COCO) @$256$ w/imprv & R-$101$ & $231.9$ & $56.2$ & $\mathbf{45.1}$\\
    \noalign{\smallskip}\hline\noalign{\smallskip}
    FPN-FCOS @$256$ & X-$64$x$4$d-$101$ & $371.2$ & $89.6$ & $43.2$ \\
    NAS-FCOS (VOC) @$128$-$256$ & X-$64$x$4$d-$101$ & $\mathbf{361.6}$ & $\mathbf{89.4}$ & $\mathbf{44.5}$\\
    NAS-FCOS (COCO) @$256$ & X-$64$x$4$d-$101$ & $368.7$ & $94.8$ & $44.4$\\
    \noalign{\smallskip}\hline\noalign{\smallskip}
    FPN-FCOS @$256$ w/improvements & X-$64$x$4$d-$101$ & $371.2$ & $89.6$ & $44.7$ \\    
    NAS-FCOS (VOC) @$128$-$256$ w/imprv & X-$64$x$4$d-$101$ & $\mathbf{361.6}$ & $\mathbf{89.4}$ & $46.1$\\
    NAS-FCOS (COCO) @$256$ w/imprv & X-$64$x$4$d-$101$ & $368.7$ & $94.8$ & $\mathbf{46.7}$ \\
    \noalign{\smallskip}\hline\noalign{\smallskip}
    FPN-FCOS @$256$ w/imprv* & X-$32$x$8$d-$101$ & $1218.9$ & $102.2$ & $49.1$\\
    NAS-FCOS (COCO) @$256$ w/imprv* & X-$64$x$4$d-$101$ & $\mathbf{1202.8}$ & $\mathbf{101.0}$ & $49.9$\\
    NAS-FCOS (COCO) @$256$ w/imprv* & X-$32$x$8$d-$101$ & $1206.3$ & $106.4$ & $\mathbf{50.0}$\\
    \noalign{\smallskip}
    \hline
    \end{tabular}
}
    \smallskip
    \caption{Results on test-dev set of COCO after full training. R-$50$ and R-$101$ represents ResNet backbones and X-$64$x$4$d-$101$ represents ResNeXt-$101$ ($64\times4$d). All networks share the same input image resolution. FLOPs and parameters are being measured on $1088\times800$, which is the median of the input size on COCO. For RetinaNet and FCOS, we use official models provided by the authors. For our NAS-FCOS, @$128$ and @$256$ means that the decoder channel width is $128$ and $256$ respectively. @$128$-$256$ is the decoder with $128$ FPN width and $256$ head width. The same improving tricks used on the newest FCOS version are used in our model for fair comparison. {*} means using multi-scale test and utilizing backbone enhanced with deformable convolutions~\cite{zhu2018deformable}.
	\label{table:det}}
\end{table*}

\subsubsection{Full Training Phase}
\label{sec:14}

In this phase, we fully train the searched models on the complete MS-COCO training dataset and select the best one by evaluating them on the MS-COCO validation images.  
Note that our training configurations are exactly the same as those in FCOS for fair comparison. Input images are resized to short size $800$ and the maximum long side is set to be $1333$. The models are trained using $4$ V100 GPUs with batch size $16$ for $90$K iterations. The initial learning rate is $0.01$ and reduces to one tenth at the $60$K-th and $80$K-th iterations. The improving tricks are applied only on the final model (w/imprv).

\begin{figure}[h!]
\centering
\includegraphics[width=0.46\textwidth]{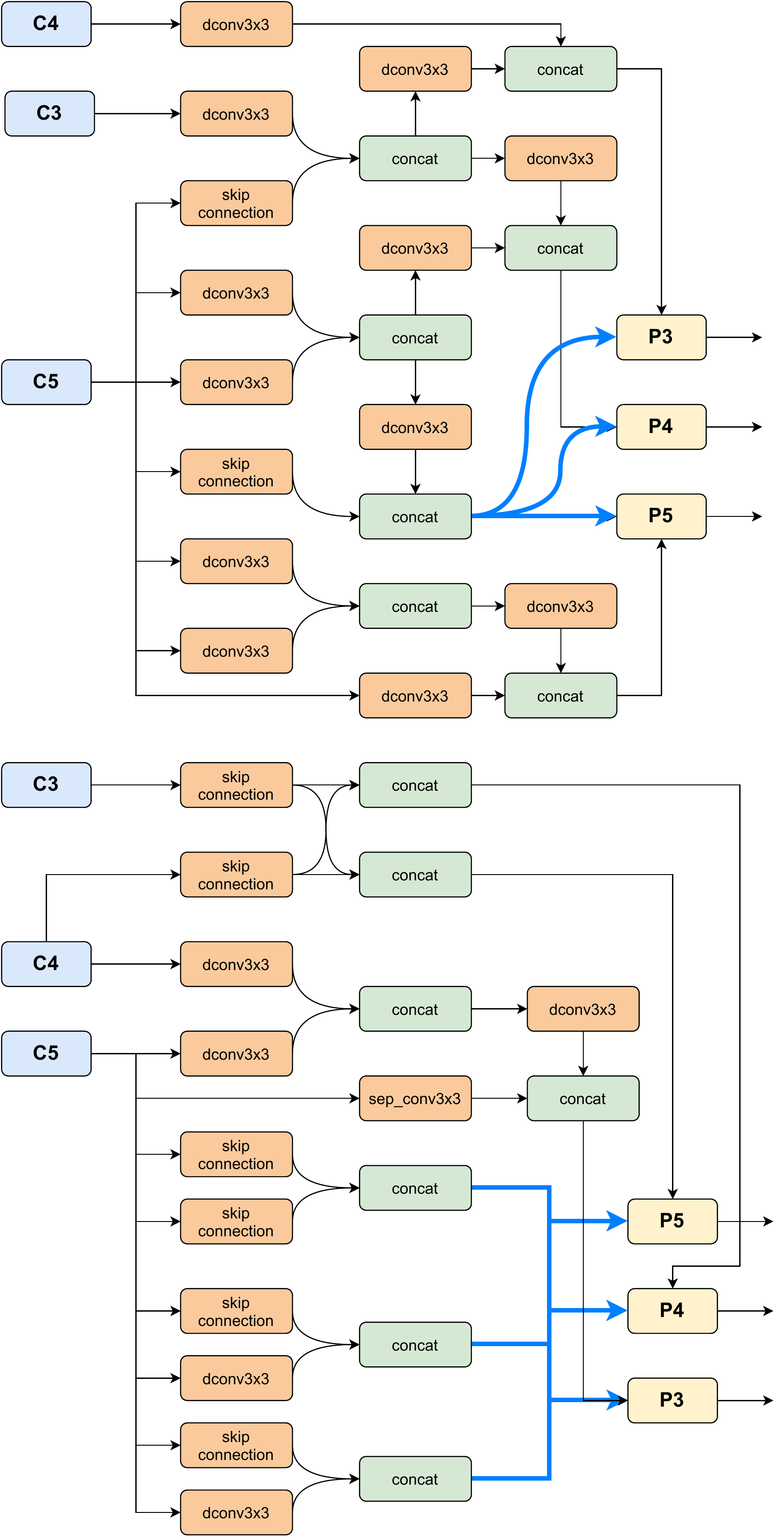}
\caption{
    Visualizations of the discovered FPN structures. Note that the top structure is based on the VOC proxy task and the bottom is on COCO.
}
\label{fig:fpn_structure}
\end{figure}

\begin{figure}[h!]
\centering
\includegraphics[width=0.5\textwidth]{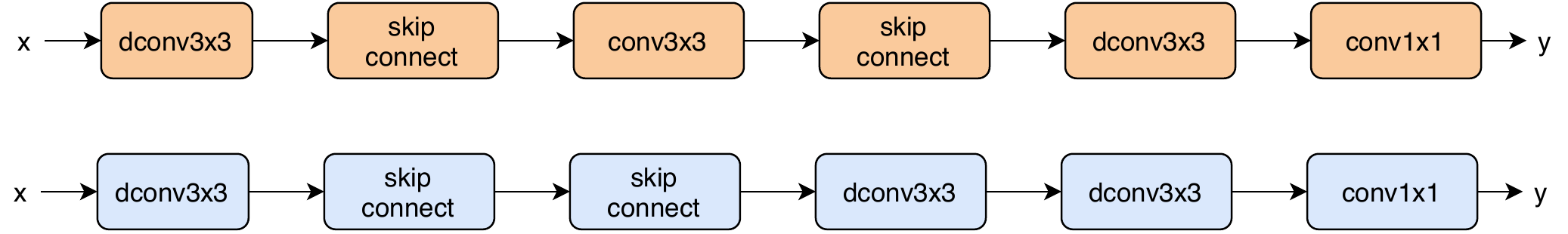}
\caption{Visualizations of the discovered head structures. The upper and lower structures correspond to the VOC and COCO proxy tasks respectively.}
\label{fig:head_structure}
\end{figure}

\subsection{Search Results}
\label{sec:15}

The best FPN and head structures based on both proxy tasks are illustrated in Fig.~\ref{fig:fpn_structure} and Fig.~\ref{fig:head_structure} respectively. As for the VOC-based FPN structure, the controller identifies that deformable convolution and concatenation are the best performing operations for unary and aggregation respectively.
And for the VOC-based head, we can see that the controller chooses to use $4$ operations (with two skip connections), rather than the maximum allowed $6$ operations. Note that the discovered ``dconv + $1$x$1$ conv'' unit structure achieves a good trade-off between accuracy and FLOPs.
Compared with the original head, our searched VOC-based head has fewer FLOPs/Params 
($79.24$G vs.\ $89.16$G, $3.41$M vs.\ $4.92$M) 
and significantly better performance (AP $38.7\%$ vs.\ $37.4\%$). 
Compared with the VOC-based one, the structure based on the COCO proxy task tends to choose fewer deformable convs ($5$ in COCO vs $12$ in VOC).

We use the searched decoder together with either light-weight backbones 
such as MobileNet-V$2$~\cite{sandler2018mobilenetv2} 
or 
more powerful backbones such as ResNet-$101$~\cite{he2016identity} and ResNeXt-$101$~\cite{xie2016aggregated}. 
To balance the performance and efficiency, we implement three decoders with different computation budgets: one with a feature dimension of $128$ (@$128$), one with $256$ (@$256$), and another with FPN channel width $128$ and prediction head $256$ (@$128$-$256$). The results on the COCO test-dev with the short side being $800$ are shown in Table~\ref{table:det}. 

\begin{figure}[t!]
\centering
\includegraphics[width=0.4\textwidth]{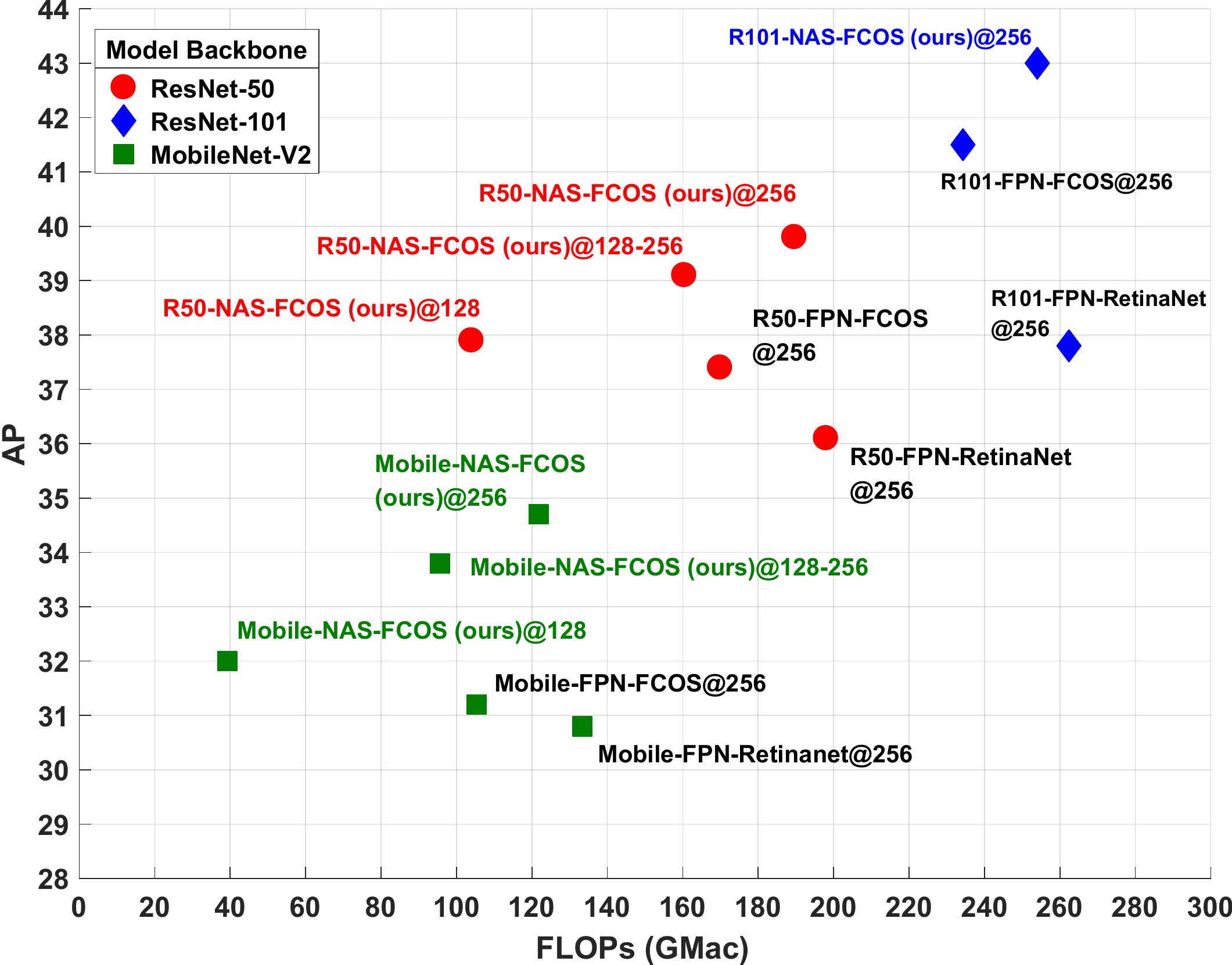}
\caption{Diagram of the relationship between FLOPs and AP with different backbones. Points of different shapes represent different backbones. NAS-FCOS@$128$ has a slight increase in precision which also gains the advantage of computation quantity. One with $256$ channels obtains the highest precision with more computation complexity. Using FPN channel width $128$ and prediction head $256$ (@$128$-$256$) offers a trade-off. Note that the searched decoder in this figure is obtained based on the VOC proxy task.
}
\label{fig:flops}
\end{figure}

\begin{figure}[t!]
\centering
\includegraphics[width=0.4\textwidth]{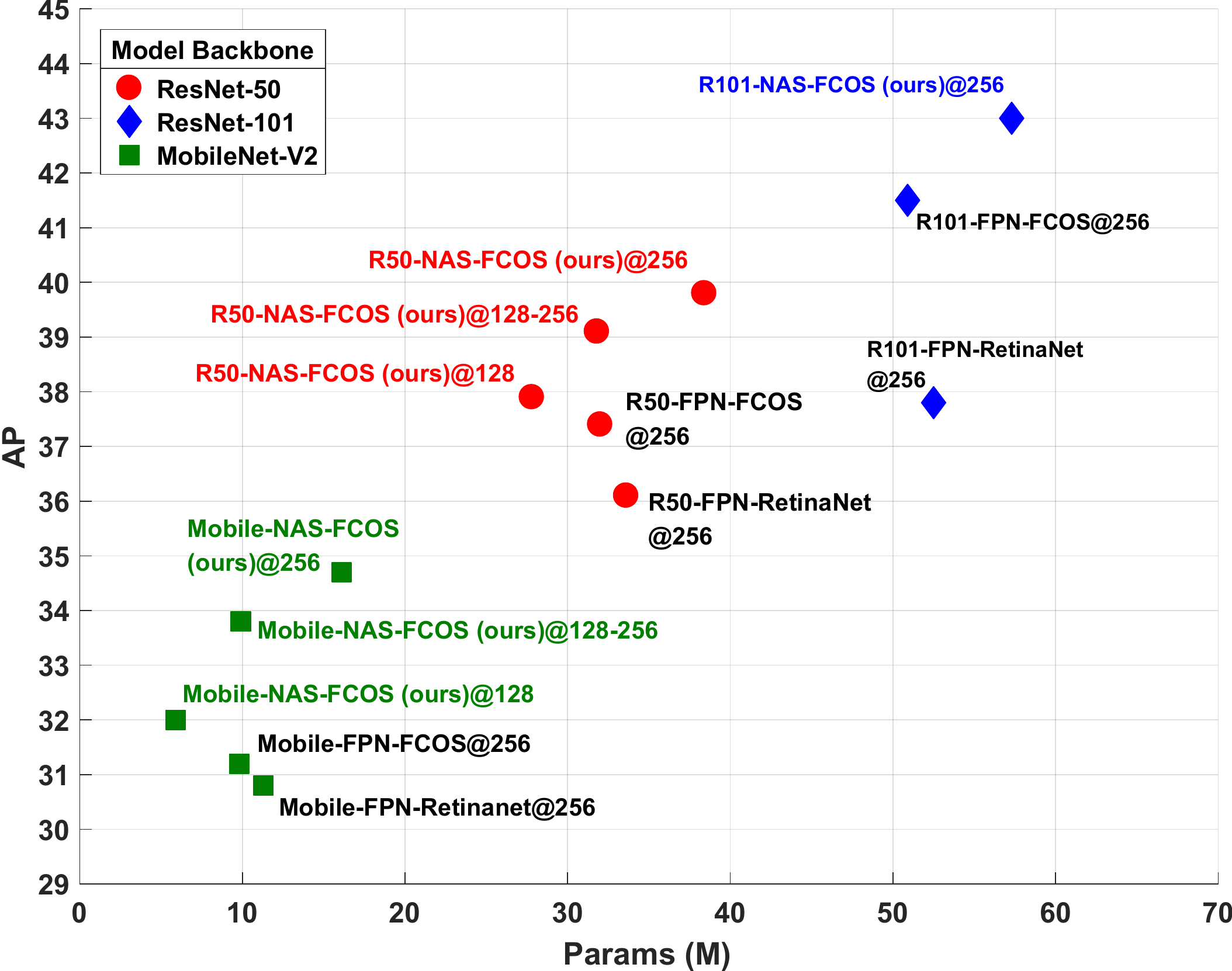}
\caption{Diagram of the relationship between parameters and AP with different backbones. Adjusting the number of channels in the FPN structure and head helps to achieve a balance between accuracy and parameters. Note that the searched decoder in this figure is obtained based on the VOC proxy task.}
\label{fig:params}
\end{figure}

With the VOC-based proxy task, the searched decoder with a feature dimension of $256$ (@$256$) surpasses its FCOS counterpart by $1.5\%$ to $3.5\%$ points in AP under different backbones. The one with $128$ channels (@$128$) has significantly reduced parameters and calculation, making it more suitable for resource-constrained environments. In particular, our searched model with $128$ channels and MobileNetV2 backbone surpasses the original FCOS with the same backbone by $0.8\%$ AP points with only $1/3$ FLOPS.
The third type of decoder (@$128$-$256$) achieves a good balance between accuracy and parameters. 
Note that our searched model outperforms the strongest FCOS variant by $1.4\%$ AP points ($46.1\%$ vs $44.7\%$) with slightly smaller FLOPs and Params. The comparison of FLOPs and the number of parameters with other models is illustrated in Fig.~\ref{fig:flops} and Fig.~\ref{fig:params} respectively.

With the COCO-based proxy task, we mainly focus on the searched decoder with a feature dimension of $256$ (@$256$). Compared with the VOC version, decoders driven by the carefully-selected COCO subset use fewer FLOPs and parameters to achieve comparable or better results. Since NAS is learning the distribution of data to some extent, we attribute the improvement to the COCO sub-set which will narrow the gap with COCO itself. For a fairer comparison, we repeat each layer in the decoder twice to make it close to the baseline FLOPs. With the light-weight backbone MobileNetV$2$, the COCO-based decoder outperforms its FCOS counterpart by $5.4\%$ AP. For ResNet-$50$, it further improves $0.5\%$ AP compared with the best VOC-based decoder with less computation cost. With the same improving tricks used on FCOS, our COCO-based model under ResNeXt-$101$-$32$x$8$d achieves $50.0\%$ AP to last.

\begin{figure}[t!]
\centering
\includegraphics[width=0.45\textwidth]{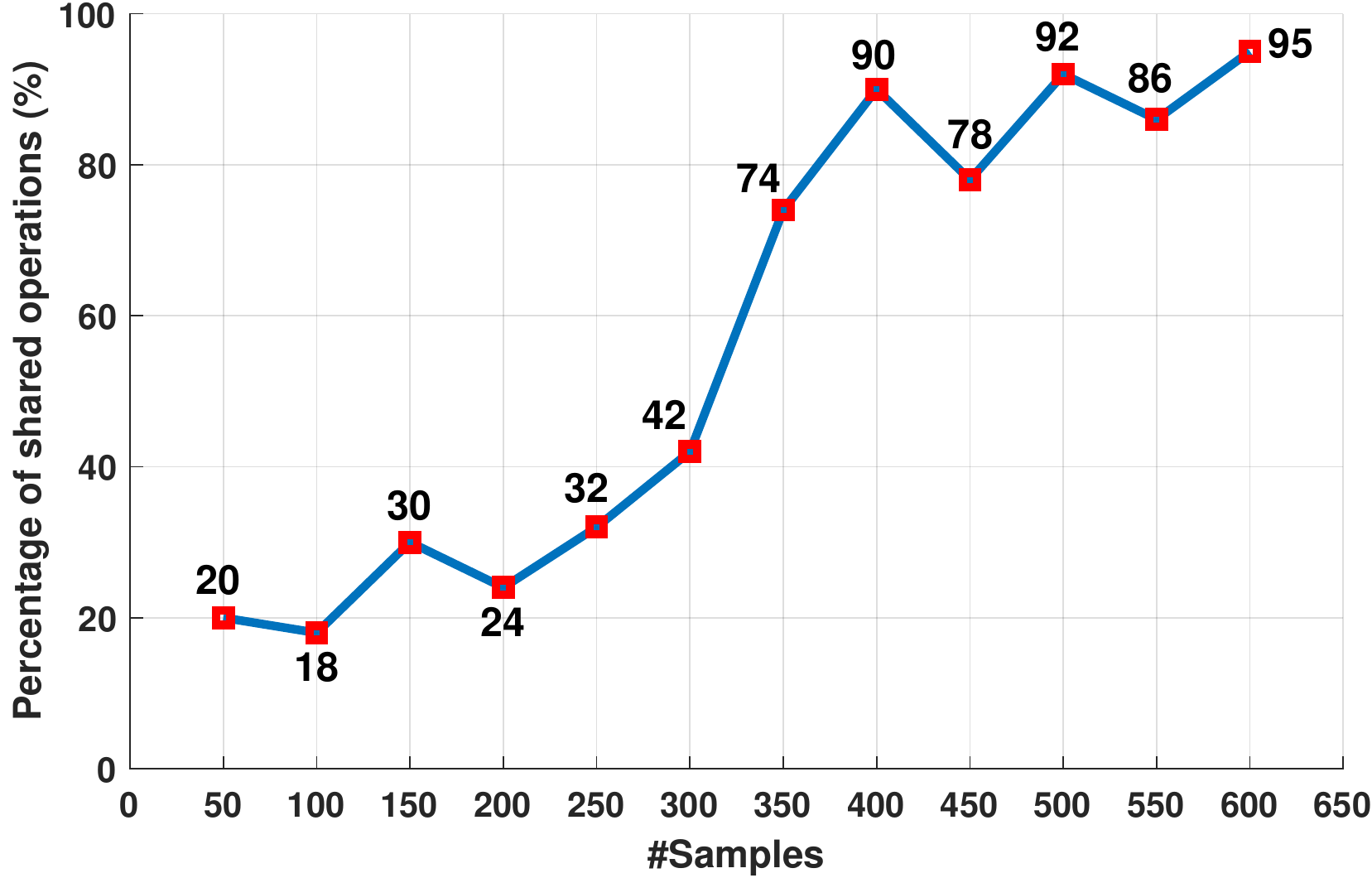}
\caption{
Change in head-level weight sharing. We compute the proportion of shared features within the structures every 50 samples as the search progresses.
}
\label{fig:share_weights}
\end{figure}

\begin{figure}[t!]
\centering
\includegraphics[width=0.45\textwidth]{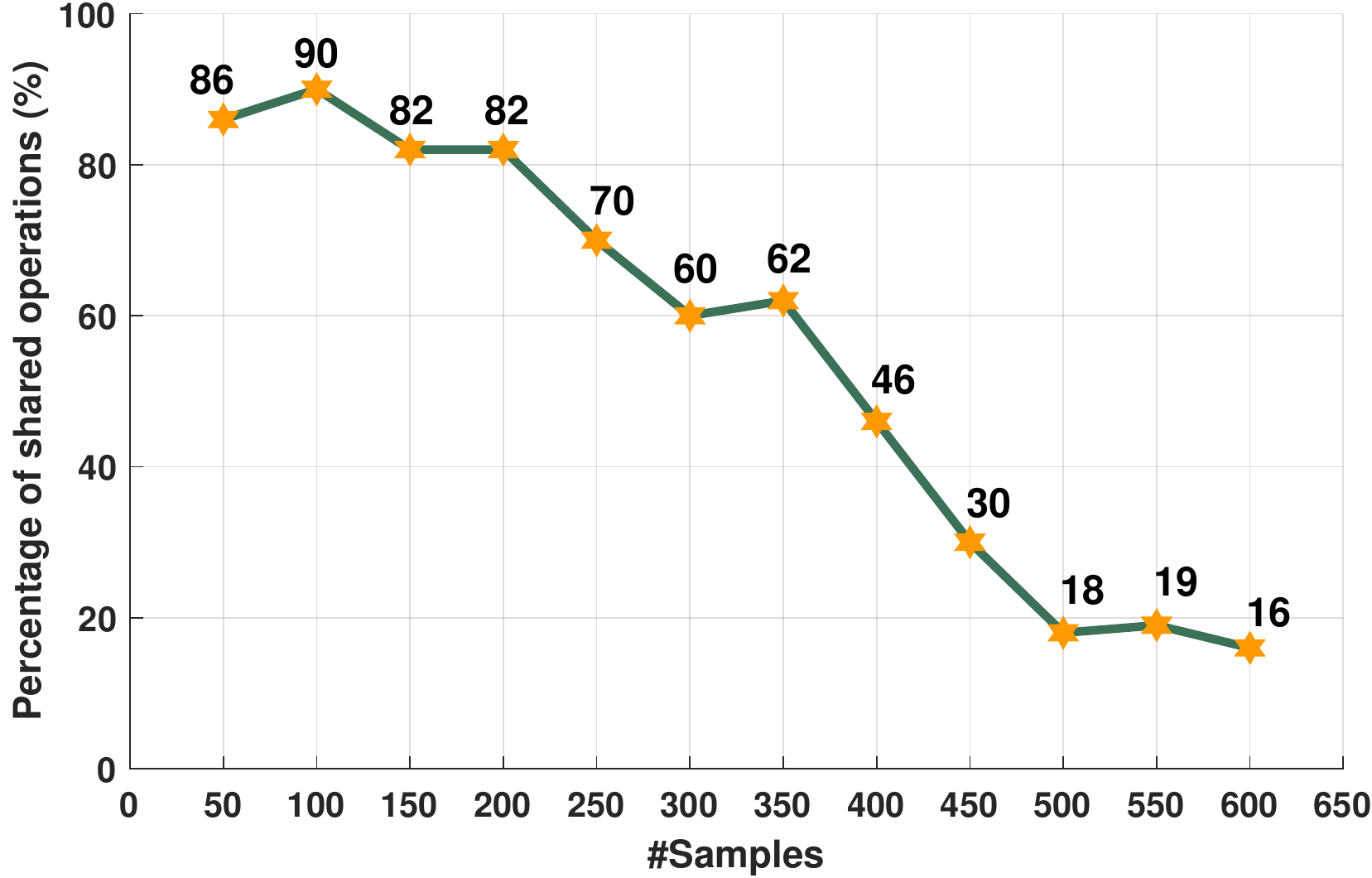}
\caption{
Change in branch-level weight sharing. As the curve shows, different branches tend to choose different features.
}
\label{fig:branch_level_share_weights}
\end{figure}

\begin{table}[t!]
\centering
\scalebox{0.90}{
    \begin{tabular}{l|c|c}
    \hline\noalign{\smallskip}
    Arch & FLOPs (G) & AP \\
    \noalign{\smallskip}\hline\noalign{\smallskip}
    FPN-RetinaNet & $198.0$ & $36.1$ \\
    NAS-RetinaNet & $195.6$ & $39.5$ \\
    \noalign{\smallskip}\hline\noalign{\smallskip}
    FPN-FCOS & $169.9$ & $37.4$ \\
    NAS-FCOS & $167.5$ & $\mathbf{40.3}$ \\
    \noalign{\smallskip}\hline
    \end{tabular}
}
    \smallskip
    \caption{The results of the search structure equipped with two different detectors. 
    Note that the structure obtained by the COCO proxy task outperforms the corresponding baselines on both two types of detectors, showing that our methods possess good transfer ability.
	\label{table:nas-retinanet}
	}
\end{table}

In order to understand the importance of weight sharing in the head, we add the number of layers shared by weights both at the head-level and branch-level as an object of the search. Fig.~\ref{fig:share_weights} and Fig.~\ref{fig:branch_level_share_weights} show trend graphs of the above two mechanisms during search respectively. We set $50$ structures as a statistical cycle. 
As the search deepens, the percentage of shared operations increases in head-level, indicating that on the multi-scale detection model, head-level weight sharing is a necessity. However, at the branch-level, models tend to choose independent weights, showing that two branches need different features to conduct their classification and regression tasks.

\begin{table*}[t!]
\centering
\scalebox{0.80}{
    \begin{tabular}{l|c|c|c|c}
    \hline\noalign{\smallskip}
    Arch & FLOPs (G) & Search Cost (GPU-day) & Searched Archs & AP\\
    \noalign{\smallskip}\hline\noalign{\smallskip}
    NAS-FPN @$256$ R-$50$ & \textgreater$325.0$ & $333\times$\#TPUs & $17000$ & \textless$38.0$\\
    NAS-FPN $7$@$256$ R-$50$ & $1125.5$ & $333\times$\#TPUs & $17000$ & $44.8$\\
    DetNAS-FPN-Faster & - & $44$ & $2200$ & $40.2$\\
    DetNAS-RetinaNet & - & $44$ & $2200$ & $33.3$\\
    \noalign{\smallskip}\hline\noalign{\smallskip}
    NAS-FCOS (COCO) @$256$ R-$50$ & $\mathbf{167.5}$ & $\mathbf{28}$ & $3000$ & $40.3$\\
    NAS-FCOS (COCO) @$256$ X-$64$x$4$d-$101$  & $368.7$ & $\mathbf{28}$ & $3000$ & $\mathbf{46.7}$\\
    \noalign{\smallskip}\hline
    \end{tabular}
}
    \smallskip
    \caption{ Comparison with other NAS methods.
    For NAS-FPN,
    the input size is $1280\times1280$ and
    the search cost should be timed by their number of TPUs used to train each architecture. 
    Note that 
    the FLOPs and AP of NAS-FPN @$256$ here are from Figure $11$ in NAS-FPN~\cite{ghiasi2019fpn},
    and NAS-FPN $7$@$256$ stacks the searched FPN structure $7$ times. 
    The input images are resized such that their shorter size is $800$ pixels in DetNASNet~\cite{chen2019detnas} and our models. 
	\label{table:nas}}
\end{table*}

Besides the typical anchor-free detector FCOS, it is still interesting to see how anchor-based models will benefit from the above proposed method. Since the entire framework does not rely on the specific components of FCOS, we directly equip the searched structure with the anchor-based detector RetinaNet. Table~\ref{table:nas-retinanet} gives the results and it outperforms more than $3.0\%$ AP compared with the origin FPN baseline under ResNet-$50$ with the COCO proxy task, showing that the anchor-based model can also benefit from our method.

We also demonstrate the comparison with other NAS methods for object detection in Table~\ref{table:nas}.
Our method is able to search for twice more architectures than DetNAS~\cite{chen2019detnas} per GPU-day. Note that the AP of NAS-FPN~\cite{ghiasi2019fpn} is achieved by stacking the searched FPN $7$ times, while we do not stack our searched FPN.
Our model with ResNeXt-101 ($64$x$4$d) as backbone outperforms NAS-FPN by $1.3\%$ AP points while using only $1/3$ FLOPs and less calculation cost.

\begin{figure}[t!]
\centering
\includegraphics[width=0.38\textwidth]{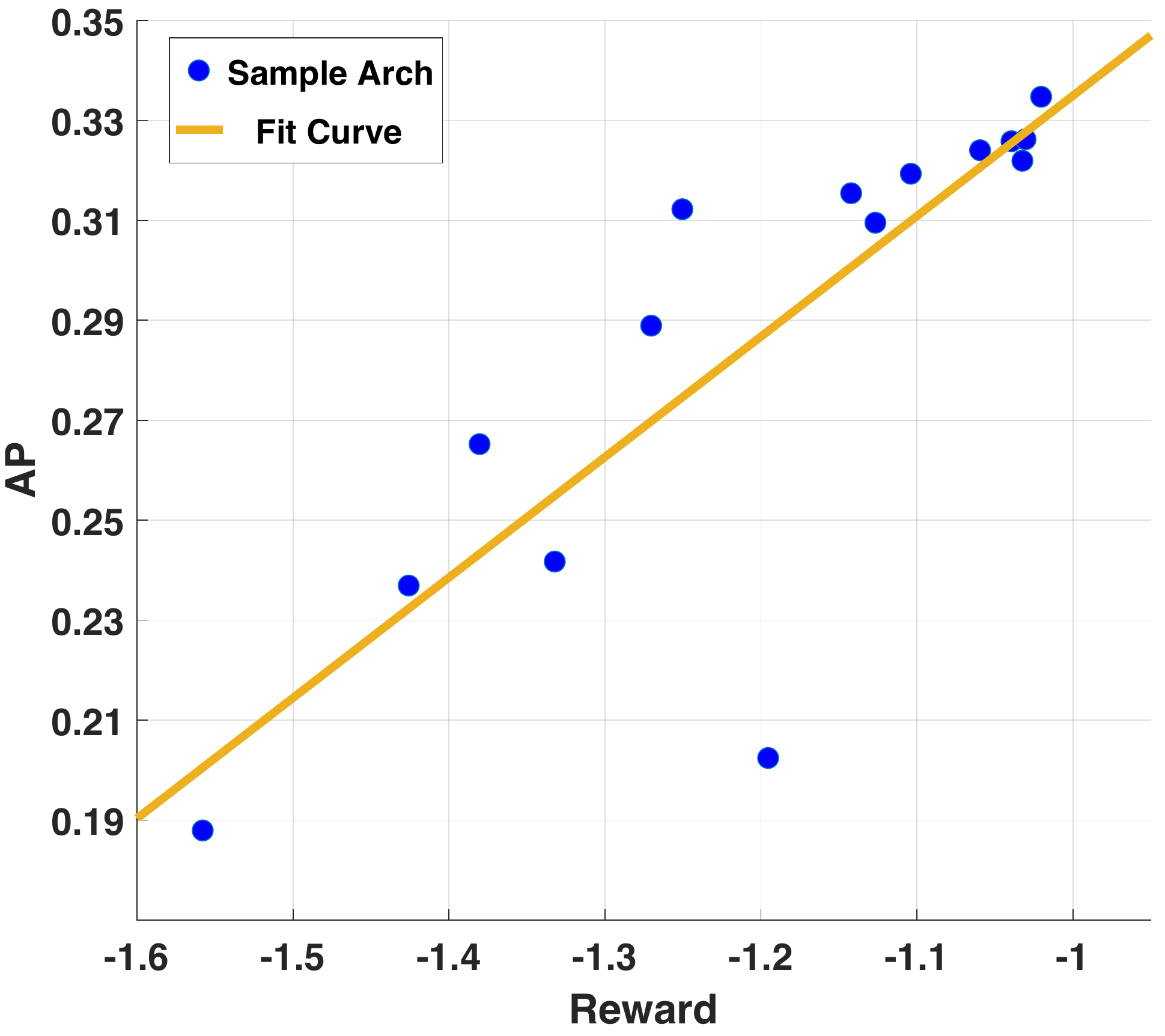}
\caption{Correlation between the search reward obtained on the VOC meta-val dataset and the AP evaluated on the COCO-val.}
\label{fig:correlation}
\end{figure}

We further measure the correlation between rewards obtained during the search process with the proxy data\-se\-t and APs attained by the same architectures trained on C\-O\-CO.
Specifically, we randomly sample $15$ architectures from all the searched structures trained on COCO with batch size $16$. Since full training on COCO is time-consuming, we reduce the iterations to $60$K. The model is then evaluated on the COCO $2017$ validation set. As visible in Fig.~\ref{fig:correlation}, there is a strong correlation between search rewards and APs obtained from COCO ($0.918$ for Spearman and $0.830$ for Pearson coefficients respectively). Poor- and well-performing architectures can be distinguished by the  rewards on the proxy task very well.

\begin{figure}[b!]
\centering
\includegraphics[width=0.438\textwidth]{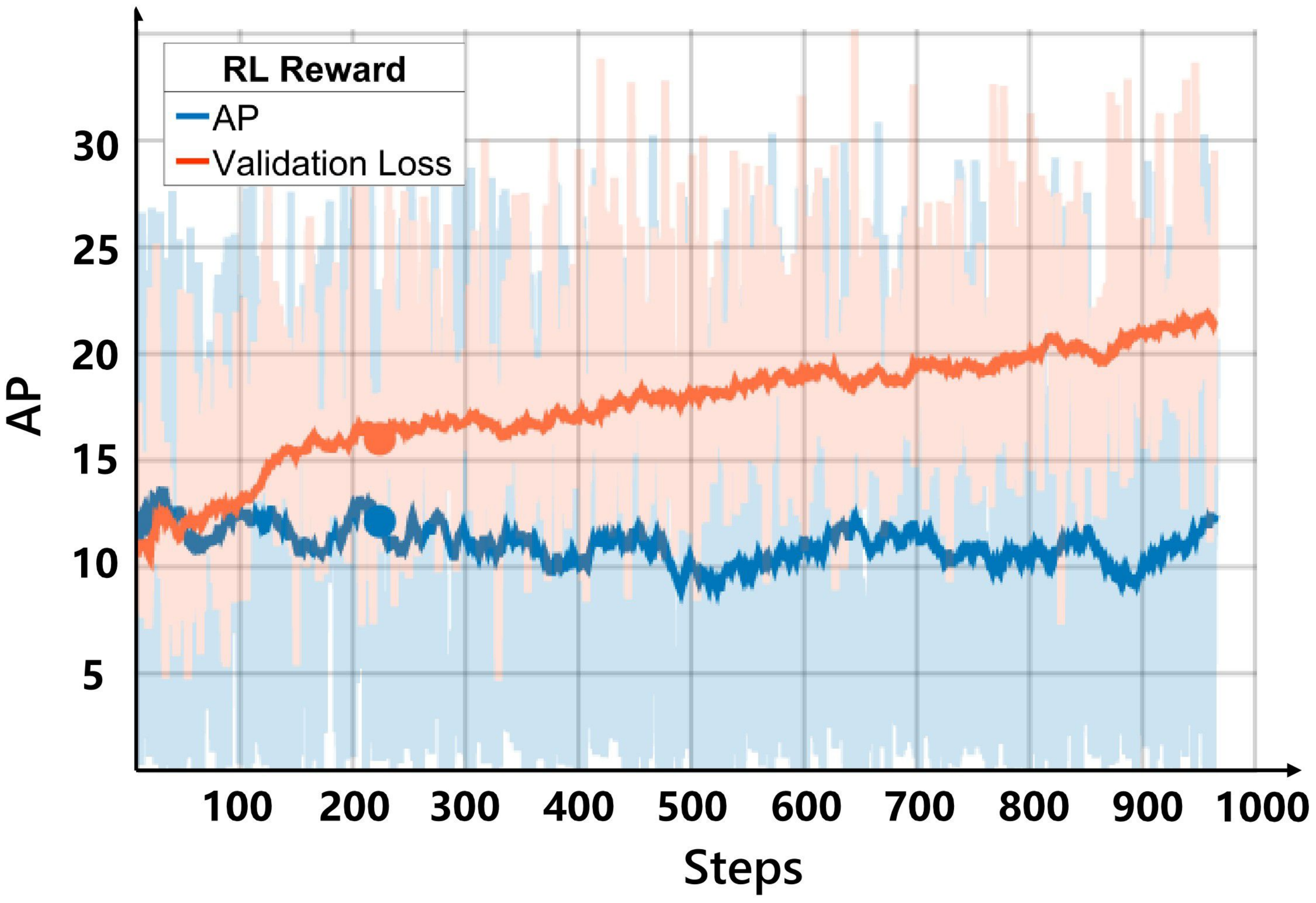}
\caption{Comparison of two different RL reward designs. The vertical axis represents AP obtained from the proxy task on the validation dataset.}
\label{fig:metric}
\end{figure}

\subsection{Ablation Study}
\label{sec:16}

\subsubsection{Design of Reinforcement Learning Reward}
\label{sec:17}
As we discussed above, it is common to use widely accepted indicators as rewards for specific tasks in the search, such as mIOU for segmentation and AP for object detection. However, we found that using AP as reward did not show a clear upward trend in short-term search rounds (blue curve in Fig.~\ref{fig:metric}). We further analyze the possible reason to be that the controller tries to learn a mapping from the decoder to the reward while the calculation of AP itself is complicated, which makes it difficult to learn this mapping within a limited number of iterations. In comparison, we clearly see the increase of AP with the validation loss as RL rewards (red curve in Fig.~\ref{fig:metric}).

\subsubsection{Effectiveness of Search Space}
\label{sec:18}
To further discuss the impact of the search spaces $f$ and $h$, we design three experiments for verification. One is to search $f$ with the original head being fixed, one is to search $h$ with the original FPN being fixed and another is to search the entire decoder ($f$+$h$). As shown in Table~\ref{table:effective}, it turns out that searching $f$ brings slightly more benefits than searching $h$ only. And our progressive search which combines both $f$ and $h$ achieves a better result.

\begin{table}[h!]
\centering
\scalebox{0.9}{
    \begin{tabular}{c|c|c}
    \hline\noalign{\smallskip}
    Decoder & Search Space & AP\\
    \noalign{\smallskip}\hline\noalign{\smallskip}
    FPN-FCOS @$256$ & - & $37.4$\\
    \noalign{\smallskip}\hline\noalign{\smallskip}
    NAS-FCOS (VOC) @$256$ & $h$ only & $38.7$\\
    NAS-FCOS (VOC) @$256$ & $f$ only & $38.9$\\
    NAS-FCOS (VOC) @$256$ & $f$ + $h$ & $\mathbf{39.8}$ \\
    \noalign{\smallskip}\hline\noalign{\smallskip}
    \end{tabular}
}
    \caption{Comparisons between APs obtained under different search space with ResNet-50 backbone.
	\label{table:effective}}
\end{table}

\subsubsection{Impact of Deformable Convolution}
\label{sec:19}
As aforementioned, deformable convolutions are included in the set of candidate operations for both $f$ and $h$, which are able to adapt to the geometric variations of objects. As shown in Table~\ref{table:deformconvstudy}, for fair comparison, we replace the whole standard $3\times3$ convolutions with deformable $3\times3$ convolutions in FPN and head of the original FCOS. Specifically, when the search space only includes $f$, the replaced deformable convolutions are repeated twice. However, the replacement is one-to-one when considering both $f$ and $h$. The above difference is to make the calculation nearly equal to our searched model (under the VOC setting). The new models are therefore called Deform-FCOS. It turns out that our NAS-FCOS model still achieves better performance (AP $= 39.8\%$ under VOC setting) than the Deform-FCOS model (AP $= 38.7\%$ replaced both in $f$ and $h$) under this circumstance. The result demonstrates that despite the candidate operations, connections between features and the design of proxy tasks also contribute to the final performance.

\begin{table}[h!]
\centering
\scalebox{0.9}{
    \begin{tabular}{l|c|c}
    \hline\noalign{\smallskip}
    Decoder & Search Space & AP \\
    \noalign{\smallskip}\hline\noalign{\smallskip}
    FCOS @$256$ & -  & $37.4$ \\
    \noalign{\smallskip}\hline\noalign{\smallskip}
    Deform-FCOS @$256$ & $f$ only  & $38.4$ \\
    NAS-FCOS (VOC) @$256$ & $f$ only & $\mathbf{38.9}$ \\
    \noalign{\smallskip}\hline\noalign{\smallskip}
    Deform-FCOS @$256$ & $f$ + $h$ & $38.7$ \\
    NAS-FCOS (VOC) @$256$ & $f$ + $h$ & $\mathbf{39.8}$ \\
    \noalign{\smallskip}\hline\noalign{\smallskip}
    \end{tabular}
}
    \caption{Comparisons between APs obtained under different settings of search space and deformable convs with ResNet-$50$ backbone. Note that @$256$ represents the channel width of the decoder is $256$.
	\label{table:deformconvstudy}}
\end{table}

\section{Conclusion}
\label{sec:20}
In this paper, we have proposed to use Neural Architecture Search to further optimize the process of designing object detection networks.
It is shown in this work that top-performing detectors can be efficiently searched using carefully designed proxy tasks, search strategies and model evaluation metrics.
The experiments on COCO demonstrates the efficiency of our discovered model NAS-FCOS and its flexibility to be used with various backbone architectures.

{
    \bibliographystyle{spmpsci}
    \bibliography{template}
}

\end{document}